\documentclass[sigconf]{acmart}
\usepackage{latexsym}
\usepackage{graphicx}
\usepackage{url}
\usepackage{algorithm,algpseudocode}
\usepackage{multirow}
\usepackage{tabularx}

\AtBeginDocument{%
  \providecommand\BibTeX{{%
    \normalfont B\kern-0.5em{\scshape i\kern-0.25em b}\kern-0.8em\TeX}}}
\acmSubmissionID{123-A56-BU3}

\newcommand\BibTeX{B\textsc{ib}\TeX}

\title{Extracting Procedural Knowledge from Technical Documents}

\author{Shivali Agarwal}
\affiliation{\institution{IBM Research India}}
\email{shivaaga@in.ibm.com}

\author{Shubham Atreja}
\affiliation{\institution{GeorgiaTech}}
\email{shubham@georgia}

\author{Vikas Agarwal}
\affiliation{\institution{IBM Research India}}
\email{avikas@in.ibm.com}
\date{}

\begin{document}

\begin{abstract}
  Procedures are an important knowledge component of documents that can be leveraged by cognitive assistants for automation, question-answering or driving a conversation.  It is a challenging problem to parse big dense documents like product manuals, user guides to automatically understand which parts are talking about procedures and subsequently extract them. Most of the existing research has focused on extracting flows in given procedures or understanding the procedures in order to answer conceptual questions. Identifying and extracting multiple procedures automatically from documents of diverse formats remains a relatively less addressed problem. In this work, we cover some of this ground by -- 1) Providing insights on how structural and linguistic properties of documents can be grouped to define types of procedures, 2) Analyzing documents to extract the relevant linguistic and structural properties, and 3) Formulating procedure identification as a classification problem that leverages the features of the document derived from the above analysis. We first implemented and deployed unsupervised techniques which were used in different use cases. Based on the evaluation in different use cases, we figured out the weaknesses of the unsupervised approach. We then designed an improved version which was supervised.  We demonstrate that our technique is effective in identifying procedures from big and complex documents alike by achieving accuracy of 89\%.  
  
\end{abstract}
\maketitle

\section{Introduction}
There is a constant endeavour to empower the virtual agents, assistants and chatbots with more and more usable and actionable knowledge. This knowledge is typically present in documents like user manuals, troubleshooting documents, guides, books and so on. Traditionally, the ability to search relevant documents, or portions of it, has been the method to use this knowledge. Now, with  virtual assistants, there is a need to develop capabilities where these agents can understand documents like a human would and suggest actions. Guided troubleshooting is an example of one such use case where the knowledge of problems and solutions in the form of step-by-step procedures can enable agents to guide users in a systematic manner. Automation efforts are a step further where an agent can carry out the actions mentioned in a document. These action steps for solving a problem are mentioned in the form of procedures in knowledge documents. Hence, automatically identifying and extracting procedures from documents becomes necessary for enabling these cognitive assistants and chatbots.

In this work, we address the problem of procedure identification in documents of varying sizes, types and formats from IT domain. 
Most of the existing research has focused on extracting flows from given procedures or understanding the entities and outcomes in the procedures to answer conceptual questions. However, extracting procedures automatically from documents other than webpages remains a relatively un-addressed problem. It is a challenging problem across domains to automatically understand from documents as to which parts are talking about procedures and consequently extract them.

One of our initial thoughts was to use structural cues to spot procedures. For example, 
 technical documents like manuals and IBM Redbooks~\footnote{http://www.redbooks.ibm.com/} have lots of information expressed using enumerated lists. We did a small manual study to find how many of these lists actually express a procedure and found that only 30--40\% of these lists were procedures and rest were some other type of enumerations. 
  Moreover, we also found a big set of troubleshooting documents where the procedures were not expressed using enumerated lists, but were nested structures. This is discussed more in Section~\ref{sec:proctypes}. We concluded that relying solely on cues like enumerations is grossly insufficient. 

\begin{figure}
  \includegraphics[width=0.48\textwidth, height = 1.1in]{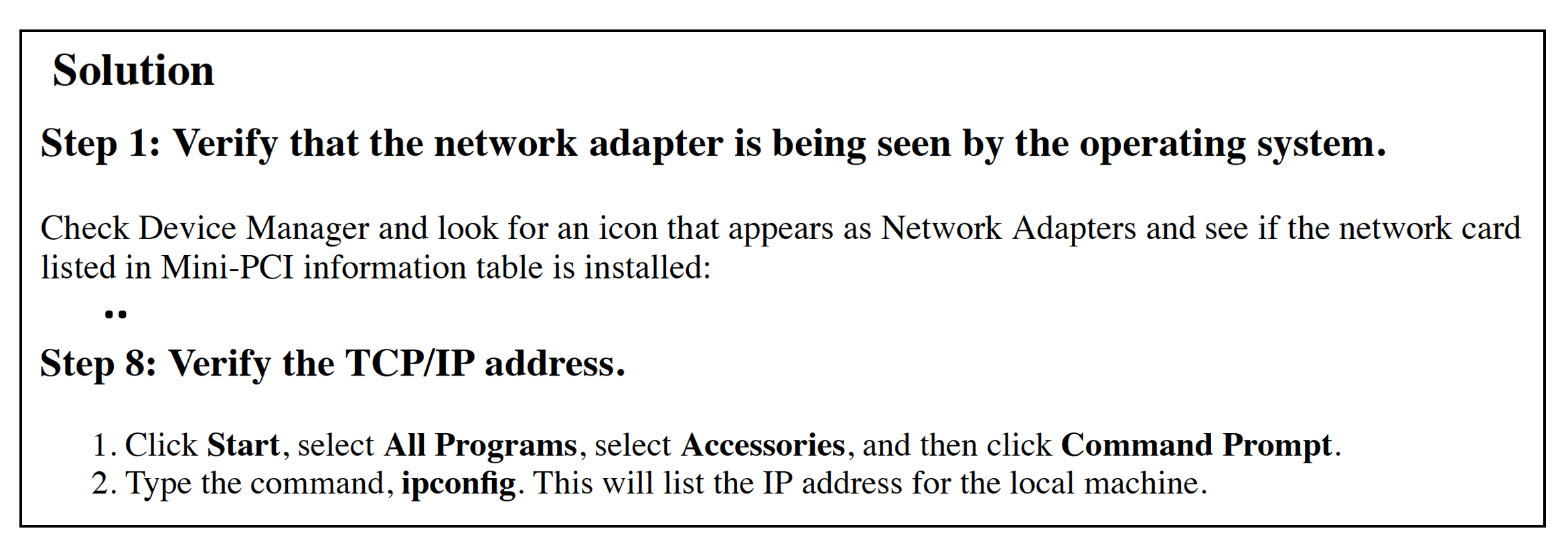}
  \includegraphics[width=0.48\textwidth, height = 1.1in]{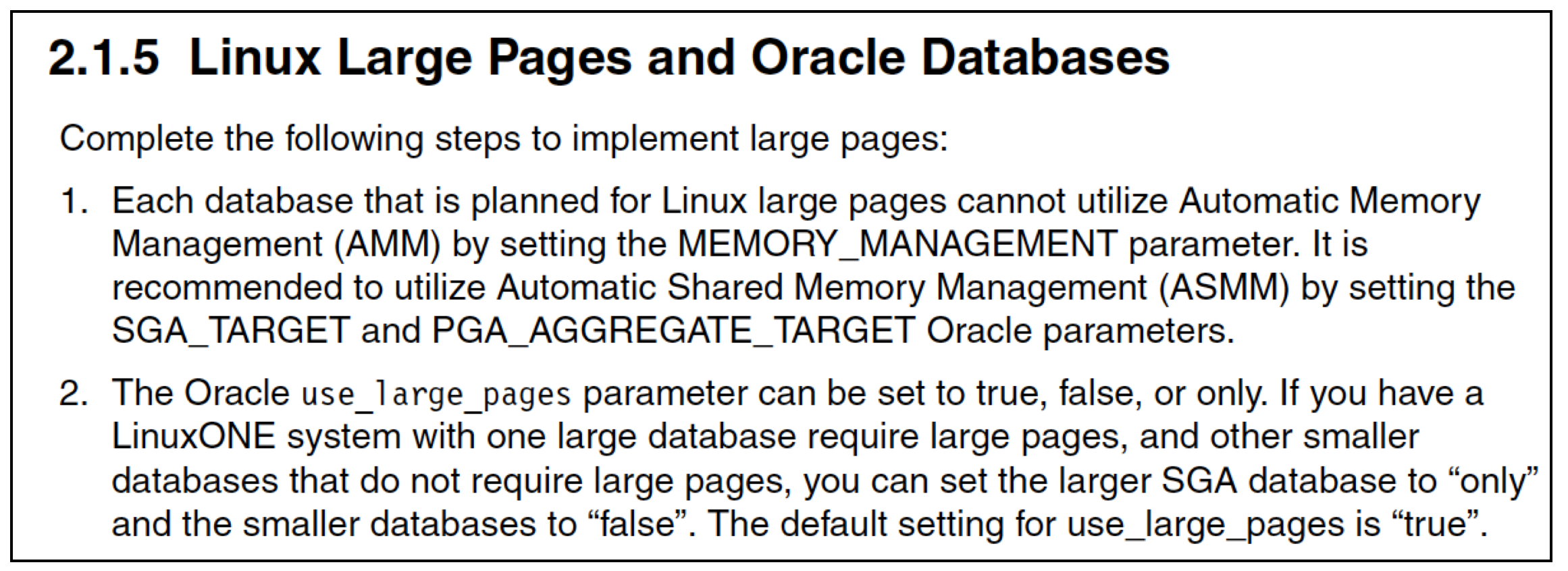}
  \caption{Procedure examples -- imperative and declarative styles}
  \label{fig:procedure-examples}
\end{figure}

One of the other main challenges is the language used to write procedures. While some documents use imperative style of writing, there are other documents that use passive, obligatory or declarative style of writing as shown in Figure~\ref{fig:procedure-examples}.
Imperative style of expression is easy to identify as actionable. Since most sentences are of declarative form, a procedure written in non-imperative language is more challenging to identify.
The dual challenge is when imperatives are used to write non-procedures as we see in many Redbooks.

Since we deal with various styles of documents, there is no standardization of how a procedure is likely to be written. In some documents, the procedure steps are written at the level of some section or sub-section heading while in others, they are part of section content. Therefore, coming up with a generic solution is very challenging. The main challenge here is to disambiguate between a procedure goal and a procedure step. In some documents, headings are goals while in others they are steps. There are also cases, where the sequence of goals denote a higher level procedure with each goal having its own procedure.

In this work, we define what are the types of procedures commonly found in documents in terms of their linguistic and structural features of expression, and then describe how to analyze a document in order to spot and extract procedures. The extraction is a two step process of classification and procedural flow creation. We use machine learning on linguistic and structural features for procedure classification. For linguistic features, we have developed a classifier for labeling the type of statements as imperative or non-imperative actionable based on state-of-the-art techniques.  In addition to classifying sentence types, we also leverage the language of discourse, that is typical of technical documents calling out for action, to identify goals and use them as cues for classification as well. We also define a method to compute relatedness or cohesion score among a sequence of sentences.  We evaluate on documents from IT domain and show the effectiveness of our approach on documents as big as books to as small as troubleshooting pages. The performance on imperative and non-imperative styles of procedures is also evaluated. The main contributions of our paper are:
1. Characterization of procedures in terms of {\it structural} and {\it linguistic} properties in order to scope the typical procedure discourse styles. 
2. Analyzing documents for specialized linguistic properties like {\it actionable statements, goals, relatedness} to name a few. 
3. Enabling identification of {\it nested procedures} using document hierarchy and knowledge propagation.
4. Formulating procedure identification as a classification problem using {\it structural and linguistic features} of document. 
5.  Providing insights into the role of linguistic and structural features in precision and recall. The evaluation of our approach on two diverse set of technical documents gives an accuracy of 88.9\%.

Given the myriad ways in which procedures can be described in a document, 
we believe our work is one of the early ones in this area that deals with complete procedure extraction, including multiple procedures, hierarchical procedures, etc., from knowledge documents given in different formats. 

Rest of the paper is organized as follows. Section~\ref{sec:defandsol} gives procedure characterization and solution overview, Section~\ref{sec:solutiondetails} describes the structural and linguistic analyses performed, and gives the classification formulation. In Section~\ref{sec:eva},  we discuss implementation details and evaluation results. 

\section{Procedure Definition and Solution Overview}
\label{sec:defandsol}
In this section, we first describe the kind of procedures that we primarily encounter in various kinds of documents and their characteristics, and then discuss our overall solution approach for identification of procedures. Once procedures are identified, we extract and represent them in a form that is easily amenable for use in different use-cases.

\subsection{Characterization of Procedures}
\label{sec:proctypes}
The concept of procedures expressed in natural language that we address in this work is formalized below.\\
A {\it procedure} is a sequence of {\it actionable steps} to achieve a {\it goal}. For example, the enumerated list items {\tt \small 1.Click Start, select ALL Programs,...Command Promp. 2. ...} under heading {\tt \small Step 8} in the first example in Figure~\ref{fig:procedure-examples} illustrates a sequence of actionable steps. Each of these steps are performing actions in a cohesive manner such that the goal, which is {\tt \small Verify the TCP/IP address} is getting realized.  A sequence of goals may also qualify to be a procedure, with each goal becoming a step, if they work together to meet a higher level goal. For example, {\tt \small Steps 1-8} in the first procedure of Figure~\ref{fig:procedure-examples} are goals forming a procedure. Hence, there can be nested procedures in the documents. A sentence that is descriptive/informational, does not qualify to be an actionable step. For example,  heading {\tt \small 2.1.5 Linux Large Pages and Oracle Databases} in second example of
Figure~\ref{fig:procedure-examples} is not an actionable step.  

Having defined what is a procedure, we now discuss how the natural language procedures that are found in documents are grouped on the basis of structural and linguistic properties of discourse. Based on the documents that we have seen, following are the prevalent structural and linguistic styles in which procedures are written.
\begin{itemize}
\item {\bf Structural Styles:} There are three styles -- \\
(S1) Steps of procedures are expressed as {\it section} or {\it sub-subsection} titles. The detailed steps are present as text under the titles. An example of this is Step 1..Step 8 in Figure~\ref{fig:procedure-examples}.\\
(S2) Steps of procedures are expressed as {\it bulleted} or {\it enumerated} lists and sub-lists. An example of this is the enumerated list under Step 8 in Figure~\ref{fig:procedure-examples}. \\
(S3) Procedures are expressed through {\it paragraph} style of writing. For example, the text under Step 1 in Figure~\ref{fig:procedure-examples}.
\item {\bf Linguistic Styles:} There are two styles -- \\
(L1) Steps expressed in {\it imperative} style of sentences e.g., Click Start, Type the command. \\
(L2) Steps expressed in {\it declarative}, {\it conditional}, or {\it passive} style of sentences e.g., the second procedure in Figure~\ref{fig:procedure-examples} has all such styles. \\
\end{itemize}

 We now provide an overview of the solution to the problem of automatically identifying and extracting procedures from documents.
\subsection{Solution Overview}
\begin{figure}
\includegraphics[width=8.0cm]{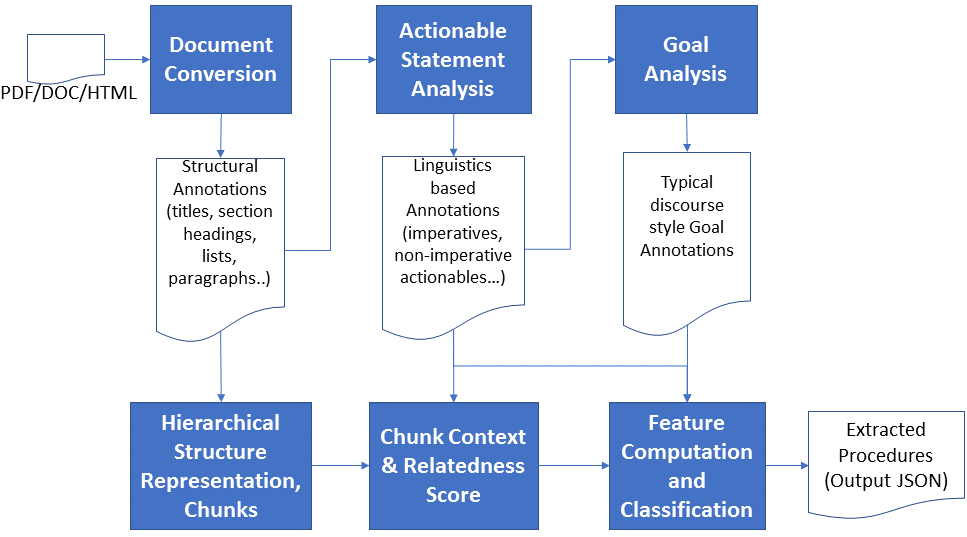}
  \caption{Component and Flow Diagram}
  \label{fig:solution}
\end{figure}

Procedure Identification problem consists of spotting the procedures present in a document text and subsequently extracting them. This problem can be broken down into multiple phases of structural and linguistic analysis as shown in Figure~\ref{fig:solution}.
The document that has to be analyzed for procedures can be of any format like doc/pdf/html. 
It is first passed through document conversion phase based on the work in \cite{CCS}, where structural annotations are obtained for the document. Essentially, the sections, paragraphs, lists, associated images, etc. are identified in this processing phase. The document can now be represented as a hierarchical tree structure. This tree structure is then traversed to identify the chunks of text which are candidates for procedures. The chunks are primarily identified using section numbering and list boundaries. 
The chunks of text are then passed through linguistic  analysis phases that provide annotations and information on constituent sentences. There are mainly three types of analysis performed as shown in Figure~\ref{fig:solution} -- actionable sentence analysis, goal analysis, and relatedness analysis. As a result, the sentences get annotated with information like imperative or conditional or non-imperative actionable, whether they denote a  goal or not. Relatedness score computes the extent of cohesive flow of information through the sentences in the chunk.  A natural language procedure can also contain branches which are manifestation of conditional sentences. In this work, we handle identification of condition-effect in conditional style sentences in order to analyze procedures that have conditionals. The actionable sentences (including both imperative and declarative) is what forms the steps of procedures. In the last phase, additional linguistic analysis is performed such as {\it discourse} based annotations, to get more linguistic cues like goals etc. Another important computation that is performed is the {\it cohesion} score of a chunk which can be thought of as a measure of relatedness. 
 As a result of these analyses, the sentences get annotated with information like whether they are imperative, conditional or non-imperative actionable, goal or not and so on.
All these structural and linguistic cues and scores are then used to compute features for each candidate chunk and a model is trained for procedure classification.

\begin{figure}
    \centering
  \includegraphics[width=0.4\textwidth, height = 2in]{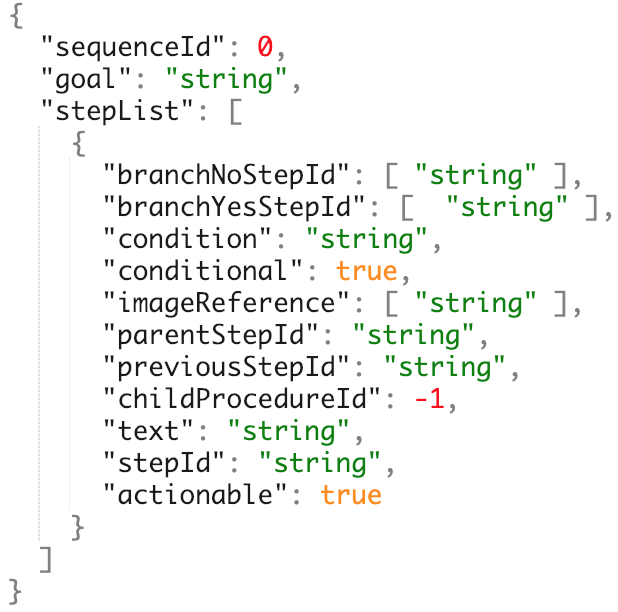}
  \caption{Sample Procedure Output in JSON Format}
  \label{fig:output-example}
\end{figure}

The output of the whole pipeline as shown in Figure~\ref{fig:solution} is a JSON array of extracted procedure objects.
Procedure extraction involves creation of a sequence of steps for chunks that are classified as procedures, along with the respective goals. 
 One such procedure model is shown in Figure \ref{fig:output-example}, where each object contains a \textit{sequenceId} as a unique identifier, a \textit{goal} statement, and a \textit{stepList} consisting of all the steps in the procedure. Each step is associated with meta-information about its linguistic role (\textit{actionable} and \textit{conditional}).  The output structure also encodes the hierarchical relations within a procedure as well as across procedures. Hierarchy within a procedure is usually in terms of steps and sub-steps (or nested lists) and is captured through the \textit{parentStepId} field. Hierarchy across procedures occurs when section headings also form a procedure. While the section heading is a step in a procedure, the text under this heading can be a procedure in itself. This can be observed in the first example in Figure \ref{fig:procedure-examples}. Such relations are captured through the \textit{childProcedureId} field associated with each step. 

The details of each of the solution components are explained in next  Section~\ref{sec:solutiondetails}.
\section{Procedure Identification Details}
\label{sec:solutiondetails}
In this section, we provide details of various components used for procedure identification as shown in Figure~\ref{fig:solution}. The analysis is done at document, chunk, and sentence levels. It shall be made clear in the exposition of the components their scope of applicability.

\subsection{Document Conversion, Hierarchical Representation and Chunks}
\label{sec:documentrep}
A document is usually organized as chapters, sections, subsections at various nested levels, and then paragraphs and lists, nested lists. A book may contain all these elements while a smaller document like troubleshooting article and webpage may contain only a subset. In order to find the procedures present, the organization of the content has to be represented using hierarchical structure like tree in order to enable identification of nested procedures. Typically, the document title is the root and each chapter title, section and subsequent subsections form hierarchical nodes as per the document. The content text, lists and paragraphs, forms the leaf nodes of this hierarchy. 

As a first step towards procedure identification, the document conversion as described in~\cite{CCS} is applied on documents to be analyzed for procedures. We obtain complete annotations of the structural organization of the content in form of titles/subtitles, lists, sublists, associated images etc. 
In the next step, the tree structure is created and  traversed to group the text into chunks using the  structural annotations in a manner that obeys the hierarchy, maintaining parent-child and sibling relationship. 
The chunks are formed as per these rules --
a) each numbered/bullet list forms a chunk,
b) section titles/headings at the same level of depth and having common parent form a chunk, and
c) paragraphs at the same depth level form chunks.

  Chunk size is total number of items in case of lists type content and total sentences in case of paragraphs and heading type content. Note that each numbered/bullet item is considered one item in list even if there are multiple sentences in that bullet. 
The {\bf context} of each chunk comes from the parent sentence or the previous sibling. The notion of previous sibling is determined by order of occurrence in the documents. These chunks are then analyzed to determine if they are procedures. The analysis is performed on each text item in the chunk as explained next. 

\subsection{Linguistic Analysis}
In this section, we describe the type of linguistic analyses performed on the document text. There are mainly three types of analyses performed -- i) if a sentence/statement is denoting an action to be performed, ii) if a chunk consists of sentences that are cohesive and related, and iii) if the discourse style of a sentence denotes a goal. 

\subsubsection{Actionable Statements}
\label{sec:actionable}
Identifying whether a given step or sentence is an actionable statement or not is an important part of overall procedure identification problem. As procedures specify steps to {\it solve} a given problem or {\it to do} a particular thing therefore, a large fraction of statements within a procedure are actionable statements i.e., they contain a perform-able action (in addition to other statements that are just informational). The steps in a procedure are generally written in an {\it imperative} style such as ``do this" or ``take this action" which directly conveys that an action is to be performed. However, they are also expressed in a {\it declarative} (or non-imperative) style, like -- ``The user enters the password", or ``The user logs into the system".

Imperative sentences are by default assumed to be actionable. We use the work in ~\cite{Coling} which in turn is built on ~\cite{DeepParsing} for classifying a sentence as an imperative or a conditional. This work also advocates how to analyze a conditional sentence for `condition' and `effect' portions. We further classify the `effect' part of the sentence as imperative or not. 

We now describe a technique to identify {\it non-imperative actionable statements} i.e., statements that describe some action to be performed but are written in a non-imperative style. We modeled the problem of actionable statement identification as a classification task. We identified two sets of features -- 1) bag-of-words features, and 2) linguistic features, that we found were relevant for identifying actionable statements.

One of the main identifiers for actionable statements are words used in the sentence. We found that actionable statements use specific action words such as {\it install}, {\it login} etc., that convey that some action needs to be performed. Some of these action words are general English language words, whereas others are domain specific words. Rather than providing a list of these action words, we let the classifier learn these from the training data.  For this, we computed the {\it tf-idf} value for each word in a sentence and used those as corresponding feature values. In-frequent words (occurring in less than 3 sentences in the training data set) were removed as they are generally proper nouns or some domain specific keywords that do not contribute in determining actionable statements. Since actionable words can vary from one domain to another, we may need to re-train our model with appropriate training data  for a different domain.

To further improve the classification accuracy, we augmented our bag-of-words features with additional features derived from linguistic analysis of sentences. Based on our analysis of procedures found in various technical documents, we identified the following linguistic features that provide cues for identifying an actionable statement. Since {\it verbs} in sentences denote the action to be performed, all these features are based on the main verb(s) in the sentence.\\
{\bf Tense:} The tense of a sentence has a distinguishing ability since actionable statements are generally expressed in present tense. Past tense is usually used to specify actions that have already happened.\\
{\bf Voice:} The voice of a sentence tells whether it is expressed actively or passively. Generally, actionable statements are written in active voice.\\
{\bf Polarity:} The polarity of a sentence has a distinguishing ability since actionable statements are usually expressed in positive sense whereas negative sense is used to indicate something that should not be done or is not related to the task at hand.\\
The implementation details and evaluation of non-imperative actionable classification are provided in Section~\ref{sec:impl-action}. 

\subsubsection{Relatedness Score}
\label{sec:relatedness}
\begin{figure}
  \includegraphics[width=0.5\textwidth, height = 1.3in]{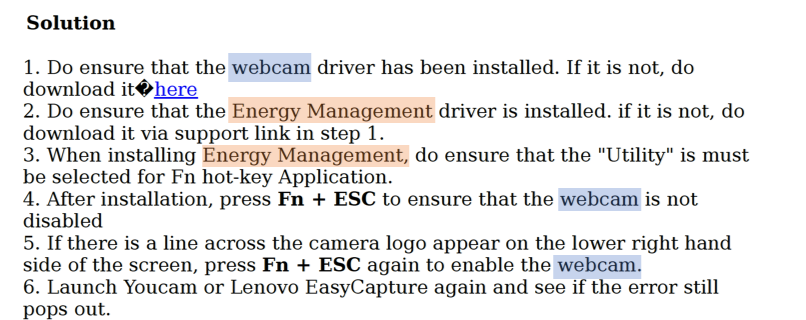}
  \caption{Entities in a procedure}
  \label{fig:relatedness-example}
\end{figure}
While all procedures have actionable statement, any set of actionable statements cannot be called a procedure. A set of statements that corresponds to a procedure must have some degree of \textit{relatedness} amongst them. This notion of \textit{relatedness} is usually captured in the sequence of actions that are performed -- in terms of entities involved in the action and the agent performing the action. As an example, Figure \ref{fig:relatedness-example} shows a basic example of how the set of entities are distributed across a sequence of statements representing a procedure.

As demonstrated in previous literature \cite{entity-coherence,graph-coherence}, entity-based representation can effectively model the local coherence of a discourse. For our task of modeling the relatedness between the statements of a procedure, we utilize an existing approach \cite{graph-coherence} that computes the discourse coherence through an unsupervised graph-based representation. It involves constructing a bipartite graph using sentences and the entities present in them as the two sets of nodes. An edge between a sentence node and an entity node is added to the graph if the entity is present in that sentence. The edge is also assigned a weight, based on the grammatical role of the entity in the sentence (either subject, object or anything else). We then apply weighted one-mode projections to the sentence node set of the bipartite graph. The projection results in a directed weighted graph denoting the connection between different sentences and the average out degree of this graph is used to model the relatedness score. For a given pair of sentence nodes, their edge weight depends on their set of common entities and the distance between them. 

Our approach differs from this existing approach in two ways. First, instead of using noun phrases as entities, we rely on a keyword extraction tool \cite{nlu} to extract the set of entities from the sentences. Second, we utilize Semantic Role Labelling (SRL) \cite{srl} while assigning weights to the edges. After generating the SRL output for a sentence, we assign weights to the edges, based on the argument (\textit{A0, A1, A2)} to which the target entity belongs. Similar to the original approach, we employ the linear weighting scheme -- the highest weight is assigned to the case where entity belongs to \textit{A0 (subject)}, followed by \textit{A1 (object)}, and then \textit{A2 (indirect object)}. The final relatedness score is a real number, where a higher value indicates a higher degree of entity-based links between the statements. 

\subsubsection{Goal Annotations}
\label{sec:discourse}
There are some discourse style cues that are useful in inferring goal from a sentence. We have used two such predominant styles. The first one is when a (sub)heading is stated using present continuous tense of the verb as an opening word e.g. ``Creating a Service Instance", then there is a high probability that this is a goal. Annotating such inferred goals act as supporting evidence for presence of a procedure in the ensuing text.  Such discourse style also qualifies to be actionable statement. Another key discourse style observed in troubleshooting documents is use of ``Method" as a prefix in (sub)headings. 
 
\begin{table*}[h]
\centering
\begin{tabularx}{\textwidth}{|l|c|l|p{11.2cm}|} \hline
   {\bf Category} & {\bf ID} & {\bf Feature Name} & {\bf Feature Description}	\\ \hline 
	\multirow{4}{*}{Actionable} & 1 & nImperatives	   & 	fraction of imperative sentences in a chunk.\\ \cline{2-4}
    	& 2 & nConditionals & fraction of conditional sentences in a chunk.	\\ \cline{2-4}
    	& 3 & nActionables & fraction of non-imperative actionables in a chunk.	\\ \cline{2-4}
    	& 4 & nEffectActionable & fraction of conditionals with actionable effect \\ \hline
	\multirow{4}{*}{Goal-based} & 5 & nDiscourseOnlyGoals & fraction of chunk items annotated with discourse style inferred goals\\ \cline{2-4}
    	& 6 & nInferredGoal & {\it propagation based computation}, fraction of chunk items that have children classified as procedure\\\cline{2-4}
    	& 7 & nNonActionableGoals & {\it propagation based computation}, fraction of nonactionable items that  have children as procedures\\\cline{2-4}
    	& 8 & ifParentIsGoal & true if parent node of chunk is annotated as a discourse style goal\\\hline
	\multirow{1}{*}{Relatedness} & 9 & relatedness & score of how items in chunk are related through entities\\\hline
	\multirow{4}{*}{Structural} & 10 & depthLevel & depth of the tree level at which chunk occurs\\\cline{2-4}
    	& 11 & chunkSize & total number of numbered/bullet items or sentences in chunk\\\cline{2-4}
    	& 12 & avgSiblingDistance & average number of items/sentences between siblings in a chunk. This is mostly relevant for chunks formed from section headings.\\\cline{2-4}
    	&13 & nAssociatedImage & fraction of chunk items that have associated image in the document\\\hline
	\multirow{2}{*}{Context-based} & 14 & contextNonProcedural & true if the context text of a chunk is indicative of chunk describing scope, properties, requirements etc. rather than a procedure\\\cline{2-4}
	    & 15 & contextProcedural & true if context text of a chunk indicates that it describes steps, flow etc.\\\hline
\end{tabularx}
\caption{Features Computed for Procedure Classification}
\label{tab:features}
\end{table*}
\subsection{Feature Computation and Procedure Classification}
\label{sec:formulation}
The structural and linguistic annotations obtained on sentences and chunks are now used for feature engineering so that the chunks can now be actually classified as a procedure or not. 
The features for classification model are computed for each chunk using the annotations at document tree level, sentence level and chunk level. 
Table~\ref{tab:features} provides the list of features used for the classification model. In this table, {\it linguistic} features are presented using the categories of {\it actionable, relatedness and goal-based}. The {\it structural} features are a category by themselves. There is another category of {\it context-based} features, which is a mix of structural (refer section~\ref{sec:documentrep}) and linguistic analysis and hence, kept separate. These features are computed at the chunk level and are thus modeled as aggregators. 

It should be noted that not all features are static in nature. Features related to goals can get updated dynamically as the classifier is run from bottom to top of the document tree. For example, if a chunk is classified as a procedure, then this information needs to propagate to the parent node in order to refine the features, in real-time, that use the knowledge of whether children are of type procedure. In our model, there are two features, namely, {\it nInferredGoal} and {\it nNonActionableGoal} that use propagation based knowledge.
We, therefore, run predicted value propagation enabled classifier on the tree from leaf chunks till root, one level at a time. 
The goal related features at one level higher are updated dynamically as the classifier is run bottom-up. This primarily enables real-time refinement of available information as new information arrives. 

\section{Implementation and Evaluation}
\label{sec:eva}
In this section, we describe the implementation details and evaluation of our procedure identification techniques including the experiments performed. 
\subsection{Actionable Statements}
\label{sec:impl-action}
For linguistic analysis, we used the Stanford natural language parsers~\cite{ref:pcfg, ref:nndep} for getting the PoS tags, constituency parse tree, and dependency parse tree. This information was then used in building the three linguistic feature extractors. For bag-of-words features we wrote our own code in Java to compute the tf-idf values.

We manually labeled 394 sentences from different types of technical documents like web pages, pdf documents, etc. as non-imperative actionable or not. These sentences were selected from procedural and non-procedural blocks within these documents. We randomly divided this into two sets - a training set of 258 sentences, and a test set consisting of 86 sentences and generated feature vectors for these. For building the SVM classifier, we used the LIBSVM~\footnote{https://www.csie.ntu.edu.tw/~cjlin/libsvm/} toolkit. We used the RBF kernel with 5-fold cross validation, parameter tuning, and feature scaling to build our model. We achieved an accuracy of 80.23\% on our test set.

\subsection{Procedure Classification}
\label{sec:proc-eval}
The framework for sections~\ref{sec:documentrep}, \ref{sec:actionable}, \ref{sec:relatedness}, \ref{sec:discourse} was implemented in Java. We used the available implementations for document conversion \cite{CCS} and imperatives and conditionals \cite{Coling}. The resulting output was a `csv' file containing the static feature set shown in Table \ref{tab:features} . The propagation and classification as described in Section~\ref{sec:formulation} was implemented using skLearn toolkit in Python. We used LinearSVC classifier with default settings.

We evaluated our proposed technique for procedure identification using two data-sets -- 1) technical support webpages a.k.a. troubleshooting documents, and 2) product documentation in the form of pdf documents i.e., IBM Redbooks. While documents in the first set are generally small in size -- describing solutions for a particular problem and containing small number of procedures, the product documentations are generally huge, up to few hundred pages and contain large number of procedural and non-procedural blocks.

We processed 5 redbooks and 12 troubleshooting documents, extracted 5021 structural annotations from document conversion, identified 693 chunks for evaluation, and tagged them as procedures or non-procedures. Out of these, we used 2 redbooks for training the classifier. These 2 redbooks had 259 chunks out of which we selected 251 for training data. This left us with 434 test candidates, 290 and 144 from the 3 redbooks and 12 troubleshooting documents respectively. Out of these 120 were actually procedures. We ensured that there were representatives of all linguistic and structural types L1,L2,S1,S2,S3 (refer section \ref{sec:proctypes}) in the training set. Importantly, we ensured presence of meaningful negative samples in both train and test data, for example, chunks that had actionable sentences but were still not procedures.

Table~\ref{tab:precision} gives the {\it accuracy}, {\it precision} and {\it recall} metrics for our procedure classification task. As can be seen, our procedure classification method was able to achieve good accuracy in identifying procedures for varied sets of documents. 

\begin{table}[ht]
\centering
\begin{tabular}{|c|c|c|} \hline
    {\bf Accuracy} & {\bf Precision} & {\bf Recall} 	\\ \hline 
	0.889	   & 	0.776  		& 0.841		\\ \hline
\end{tabular}
\caption{Procedure classification metrics}
\label{tab:precision}
\end{table}
We further performed experiments to gain insights into the role of different features that were used. This was done by selective feature elimination. We removed the feature(s), that we wanted to study the effect of selectively, from the training data and the test data and ran the classifier. The impact on accuracy metrics by removing the features corresponding to different categories (Table\ref{tab:features}) is shown in Table~\ref{tab:featureelimination}.


\begin{table}[ht]
\centering
\begin{tabular}{|l|c|c|c|} \hline
    {\bf Feature Type}&{\bf Accuracy} & {\bf Precision} & {\bf Recall} 	\\ \hline 
	Actionable &0.829	   & 	0.722  		& 0.613		\\ \hline
	Structural & 0.834 & 0.644 & 0.882 \\\hline	
	Relatedness & 0.843 & 0.654 & 0.891 \\ \hline
	Goal-based &0.822 & 0.623 & 0.890 \\\hline
	Context-based & 0.882 & 0.768 & 0.808 \\\hline
\end{tabular}
\caption{Effect of Removing Features on Classification}
\label{tab:featureelimination}
\end{table}
\noindent {\bf Discussion} Table \ref{tab:featureelimination} shows that there is a constant trade off between precision and recall when different sets of features are selectively removed. From the table, one can observe that both {\it structural} and {\it linguistic} features are important for identifying the procedures. While structural features are key to precision, the group of linguistic features play vital role in both precision and recall. Removing {\it actionable} set of features results in a huge drop in recall metrics which is as expected with decrease in precision as well.
On the other hand, removing any of {\it relatedness} or {\it goal-based} set of features results in a sharp decrease in precision accompanied with an increase in recall. The result is intuitive, as relatedness and goals are mainly used to filter out negative examples, i.e., ones that contain actionable items but do not constitute a procedure. Removing {\it structural} group of features causes sharp drop in precision. We found that that the depth level, chunk size and number of images associated play an important role in ruling out false positives.  The results also show that removing  {\it context-based features}, which are derived using basic structural and linguistic properties of document also leads to a reduction in recall, although the impact is limited compared to the other features.

\section{Related Work}
There have been several works in past in the area of procedural text, but most of them are focused on knowledge extraction from procedural and instructional text for building a knowledge base for information retrieval tasks. \citet{ref:Zhang2012AutomaticallyEP} provides a technique for extracting procedural knowledge such as action, purpose, actee, instrument, etc. from instructional texts and representing it in a machine understandable format for automation purposes. \citet{ref:Jung2010ACL} describes an approach for building a situation ontology by automatically extracting tasks, actions and their related information from how-to instructions available in {\it wikiHow} and {\it eHow} on Web. \citet{ref:Park2018LPT} have a similar aim of identifying relationships such as goals, tasks and sub-tasks, but within a procedure itself. In~\cite{ref:statechanges} the authors describe an approach for improving the predicted effects of actions in a procedure by incorporating global commonsense constraints. All these works are mainly focused on deeper understanding of procedural knowledge in terms of actions and entities involved, to build a machine understandable knowledge base for answering questions related to it. But, they do not deal with complete procedural text extraction from general documents which is the main aim of our work.

Another body of work is in the area of action statements identification from text. \citet{ref:eAssistant} describe a system called eAssistant which identifies actionable statements such as a request or a promise from email and conversational text for notifying users of important content that requires their attention. Similarly, \citet{ref:actionable} describe a method for automatically detecting actionable clauses from how-to instructions for problem solving tasks. While these work don't address the problem of complete procedural text extraction from documents, they address a small part of our overall task related to {\it step identification} that requires us to first detect whether a given statement is actionable or not. We build upon and use some of the linguistic features described in these work for actionable statement identification in Section~\ref{sec:actionable}.

There has been very little work on complete procedural text extraction from knowledge documents. The closest that we could find is \cite{ref:abhirut} where the authors describe a method for extracting procedures from web pages and further analyzing them to identify decision points for guided troubleshooting purposes. In~\cite{ref:LREC2008}, the authors provide linguistic analysis of procedures and describe a technique for extracting procedures and its components 
for answering how-to questions where the response is a complete procedure. 
While the overall goal of these works is similar to ours, they only handle simple procedures that are expressed as a list in a webpage. They do not handle more complex procedure representation such as those expressed using section/sub-section headings or a hierarchy of procedures. These are more common in technical documentation available as MS-Word or PDF documents that we are able to handle. 


\section{Conclusion}
Procedures are an important actionable component of knowledge documents that are important for variety of tasks such as enabling cognitive assistants and chatbots. 
In this paper, we described a very effective technique for automatically identifying and extracting procedures from different size, type and style of documents. We first presented a structural and linguistic characterization of procedures found in documents and then described techniques for analyzing documents both structurally and linguistically to extract these features. 
We then used classification technique built on these features that used structural and linguistic features 
to identify procedural chunks in documents. The evaluation showed that we were able to achieve good accuracy on different document types.
This paper mainly focused on documents from IT domain. As part of future work, we want to extend our technique to other domains like finance, compliance etc. and see if we can transcend the boundaries of domain and develop a generalized model. 

In this paper we mainly focused on well formed documents i.e, those where procedure steps are properly formatted as section/sub-section titles or list items. In reality, we have found many documents that are badly formatted. For example, instead of using proper list type formatting (bullets or numbers), the procedure steps are generally mentioned on separate lines using paragraph style of formatting or manually add numbering as part of step text. Moreover, there can be a mix of both formatted and un-formatted steps in a procedure. These issues create additional challenges for the already complex task. We are currently exploring techniques to correctly identify the boundaries of procedures in presence of formatting issues in the documents.



\bibliographystyle{ACM-Reference-Format}
\bibliography{acl2019}

\appendix



\end{document}